\documentclass[prb,showpacs,twocolumn]{revtex4}

\usepackage{snapshot}
\usepackage{bm}
\usepackage[usenames]{color}
\usepackage{listings}
\usepackage{amsmath,amssymb,latexsym,amsfonts,graphicx}

\begin{document}
\title{Anomaly Sequences Detection from Logs Based on Compression}
\author{Wang Nan}
\author{Han Jizhong}
\author{Fang Jinyun}
\affiliation{Institute of Computing Technology, Chinese Academy of Sciences,
Beijing 100190, China}

\begin{abstract}
Mining information from logs is an old and still active research topic. In
recent years, with the rapid emerging of cloud computing, log mining becomes
increasingly important to industry. This paper focus on one major mission of
log mining: anomaly detection, and proposes a novel method for mining abnormal
sequences from large logs. Different from previous anomaly detection systems
which based on statistics, probabilities and Markov assumption, our approach
measures the strangeness of a sequence using compression. It first trains a
grammar about normal behaviors using grammar-based compression, then measures
the information quantities and densities of questionable sequences according to
incrementation of grammar length. We have applied our approach on mining some
real bugs from fine grained execution logs. We have also tested its ability on
intrusion detection using some publicity available system call traces. The
experiments show that our method successfully selects the strange sequences
which related to bugs or attacking.
\end{abstract}

\maketitle

\section{Introduction}

\noindent From trend analysing to system tuning, log mining technique is widely
used in commercial and research area. In recent years, diagnosing systems
according to logs becomes a hot research topic because of the rapid emerging of
cloud computing systems\cite{yuan2010sherlog, yuan2011improving,
zhang2011autolog, tan2009mochi, xu2009detecting}. Problems in such systems are
always non-deterministic because they are caused by uncontrollable conditions.
Therefore, developers can watch neither the execution paths nor the
communications between different components as they used to be. The only cute
can be used are the text logs generated by buggy systems. However, the huge
size of such logs makes developers hard to deal with them.

Aiming at non-deterministic problems, many approaches are proposed.
Record-replay\cite{r2, pres, snitchaser, rebranch} is a hopeful one.
Record-replay systems record low level execution detail during running. When
debugging, they can replay the buggy execution according to those data, let
developers to check control flow and data flow of the target programs. 
Nevertheless, although recent record-replay tools can achieve low performance impact
to be suitable for deploying into production environments\cite{pres, snitchaser},
replaying 7x24 long lasting logs and manually identifying the key parts of the
execution flow is still an obstacle.

The heart of the above problems is finding unusual patterns from large data set.
This is the goal of intrusion detection. There are 2 general approaches of
intrusion detection: misuse intrusion detection (MID) and anomaly intrusion
detection (AID). MID models unusual behaviors as specific patterns and
identifies them from logs. However, MID systems are vulnerable against unknown
abnormal behaviors. This paper focus on AID, which models normal behaviors and
reports unacceptable deviations.  Anomaly detection has already been studied for
decades. The related techniques are used for detection of network intrusion and
attacking\cite{denning1987intrusion, quao2001anomaly, hyun2003anomaly,
tian2010network}. Those approaches apply probability models and machine learning
algorithms\cite{quao2001anomaly, tian2010network, hyun2003anomaly}, most of them
rely on Markov assumption. They have achieved positive results on some specific
type of logs such as system call traces\cite{quao2001anomaly, tian2010network,
warrender1999detecting}. However, although Markov assumption makes them
sensitive to unusual state transitions at low level, they are short to identify
high level misbehavior.

This paper proposes a novel anomaly detection method. Different from the above
approaches, our method doesn't rely on statistics, probabilities or Markov
assumption, and needn't complex algorithms used in machine learning. The
principle of our approach is straightforward: using compression to measure the
information quantities of sequences. Our method can be used to find some high
level abnormal behavior. To the best of our knowledge, our work is the first
attempt to utilize the relationship between information quantities and
compression in mining unusual sequences from logs.

We first introduce the principle of our approach using a simple example in
section \ref{sect:overview}, then present the detail algorithms in
section \ref{sect:detail}. Section \ref{sect:experiment} lists a set of
experiments to show the ability of our method on bug finding and intrusion
detection. Section \ref{sect:conclusion} concludes the paper.

\section{Overview}\label{sect:overview}


Our approach is inspired from following obvious fact.

When the normal behavior of a information source is known to the viewer,
she can describe another normal sequence use only a few words, but needs more words
to describe an abnormal sequence. For example, the viewer is told
that \textsl{``1234 1235 1234 1235''} are 4 normal sequences. For
a questionable sequence \textsl{``1235''}, she can describe it as \textsl{``another type II
sequence''}. The information contained in her description is only the
\textsl{``type II''}. However, for sequence \textsl{``1237''}, the most elegant
description should be \textsl{``replace the forth character of normal pattern by 7''}.
The information contained in this description is \textsl{``forth''} and
\textsl{``7''}. For sequence \textsl{``32145''}, she has to say \textsl{``a new
sequence, the first character is 3, the second character is 2 ...''}. The
information contained in this description is much more than the previous two.
Therefore, the viewer can infer that the last sequence is \textsl{``the most
strange one''}. An anomaly detection system should report the last
sequence among the three.

In computer science there is a method to ``describe a sequence'': compression.
A compression algorithm can reduce the size of a sequence. For a long sequence,
the compressed data can be thought as ``the description of the original
sequence''. It is well known that no compression algorithm is able to ultimately
reduce the size of a sequence to zero. It is also well known that a compressed
file is hard to be compressed again, entropy rate ($H(X)$) of the source data
restricts the performance of a compression algorithm.

Our approach utilizes the relationship between the information quantity and the
compressed data size. To select the \textsl{``most strange''} sequence, we can
use the following 3 steps:

\begin{itemize}

\item Training: compress a set of normal sequences, the compressed data size is
$Q_0$.

\item Evaluating: for each candidate sequence, add it into the normal set used
in training step, compress the new set. The compressed size is $Q_n$. Let $I_n
= Q_n - Q_0$.

\item Selecting: the $n$th sequence which generates the largest $Q_n$
is selected.

\end{itemize}

We use the previous example to demonstrate the above 3 steps. The compression
algorithm is \texttt{gzip}.

\doublerulesep 0.1pt
\begin{table}
\begin{footnotesize}
\caption{The size of gzipped data}
\label{tab:gzipexample}
\begin{tabular}{p{.3cm}p{2cm}p{3cm}p{.5cm}p{0.9cm}}
\hline\hline\noalign{\smallskip}
$n$ & evaluated sequence & gzipped sequence    &  $Q_n$ &  $I_n$  \\
\noalign{\smallskip}
\hline
0 &      & 1234123512341235  &   30   & 0 \\
1 & 1234 & 12341235123412351234 & 30  & 0 \\
2 & 1237 & 12341235123412351237 & 31  & 1 \\
3 & 32145& 123412351234123532145 & 35  & 5 \\
\hline\hline
\end{tabular}
\end{footnotesize}
\end{table}

The first row of table \ref{tab:gzipexample} shows the result of training.
Sequence \textsl{1234123512341235} is combined from 4 normal sequences,
gzip compresses it into 30 bytes. Following rows show the
evaluating step. 3 questionable sequences are appended then
gzipped. The incrementation of the three are 0, 1 and 5. The third step selects
\textsl{32145} as ``the most strange'' one as it generates the largest
$I_n$.

\section{Detail}\label{sect:detail}

In this section we introduce our approach in detail.

\subsection{Grammar-based compression algorithm}

Although we use gzip algorithm to explain the principle of our anomaly detection
method in section \ref{sect:overview}, gzip and other well known generic compression
algorithms are not suitable for digging anomaly sequences from execution logs
because of the following reasons:
\begin{itemize}

\item Nearly all well known compression algorithms (such as gzip, bzip2 and rar)
are based on LZ77\cite{lz77}, which uses sliding window to store recent data for
matching incoming stream and ignore previous data. Sliding window is important for
a generic compression algorithm for compressing speed. However, it eliminates
the historic knowledge about the data source, makes those data
effectless when compressing new data.

\item Generic compression algorithms compress data as byte stream. Their
alphabets are 256 possible bytes from 0x0 to 0xff. However, the unit of execution
logs is log entry. A compression algorithm with alphabet made by possible
entries can discover meaningful patterns.

\item Generic compression algorithms are unable to identify difference sequences
when training and evaluating. Sequences in execution logs have to be stick
together one by one. Some patterns will be created unintentionally across
different sequences and affect the evaluating processing. In previous example,
generic compression algorithms take pattern \textsl{``34123''} into account
although it is not a part of any sequences.

\end{itemize}

Our approach chooses a grammar-based codes as the underlying compression algorithm.
Grammar-based compression algorithms are developed recent decades as a new way to
losslessly compress data\cite{kieffer2000grammar, yang2000efficient,
nevill1997identifying, nevill1997compression}. Kieffer et
al.\cite{kieffer2000grammar} firstly published some important theorems on it.
This paper uses same symbols and terminologies to describe the algorithm.
Yang et al.\cite{yang2000efficient} presented a greedy grammar transform.
Our algorithm is based on it.
Such grammar transform is
similar to SEQUITUR\cite{nevill1997identifying, nevill1997compression}, but
generates more compact grammars. 

The idea of grammar-based compression is simple: for stream
$x$, one can represents it as a context-free grammar $G_x$ which generates
language $\{ x \}$ and takes much less space to store. Grammar-based compression
is suitable for compressing execution logs because such logs are generated by program
hierarchically and are highly structured. 

\subsubsection{Grammar transform overview}

Grammar transform converts a sequence $x$ into an admissible grammar $G_x$ that
represents $x$. An admissible grammar $G_x$ is such a grammar which guarantees
language $L(G) = \{x\}$. (The language of $G_x$ contains only $x$.) We
define $G = (V, T, P, S)$ in which
\begin{itemize}
\item $V$ is a finite nonempty set of \emph{non-terminals}.
\item $T$ is a finite nonempty set of \emph{terminals}.
\item $P$ is a finite set of \emph{production rules}. A production rule is an
expression of the form $A \to \alpha$, where $A \in V$, $\alpha \in (T \cup
V)^{+}$. ($\alpha$ is nonempty).
\item $S \in V$ is \emph{start symbol}.
\end{itemize}

Define $f_G$ to be endomorphism on $(V(G) \cup T(G))^*$ such that:
\begin{itemize}
\item $f_G(a) = a, \quad a \in T(G)$
\item $f_G(A) = \alpha, \quad A \in V(G)$ and $A \to \alpha \in P(G)$
\item $f_G(\epsilon) = \epsilon$
\item $f_G(u_1 u_2) = f_G(u_1)f_G(u_2)$
\end{itemize}

Define a family of endomorphism $\{f^k : k = 0, 1, 2, \cdots\}$:
\begin{itemize}
\item $f^{0}_{G}(x) = x$ for any $x$
\item $f^{1}_{G}(x) = f_G(x)$
\item $f^{k}_{G}(x) = f_G(f^{k-1}_{G}(x))$
\end{itemize}

Kieffer et al. showed\cite{kieffer2000grammar} that, for an admissible grammar
$G_x$, $f_{G_x}^{|V(G_x)|}(u) \in (T(G_x))^{+} $ for each $u \in (V(G_x) \cup
T(G_x))^{+}$, and $f_{G_x}^{|V(G_x)|}(G_x(S)) = x$. Informally speaking, for an
admissible grammar $G_x$, by iteratively replacing non-terminals with the right side
of corresponding production rules, every $u \in (V(G) \cup T(G))^{+}$ will
finally be translated into a string which contains only terminals.
Define a mapping
$f^{\infty}_G$ such that $f^{\infty}_G(u) = f_{G}^{|V(G)|}(u)$ for each $u
\in (V(G) \cup T(G))^{+}$. Informally speaking, $f^{\infty}_G(u)$ is the
original sequence represented by $u$.

\subsubsection{The greedy grammar transform algorithm}

The algorithm we used is based on following reduction rules (in following
description, $\alpha$ and $\beta$ represent string in $(V(G) \cup T(G))^{*}$):
\begin{enumerate}

\item For an admissible grammar $G$, if there is a non-terminal $A$ which
appears at right side of production rules only once in $P(G)$, let $A \to
\alpha$ be the production rule corresponding to $A$, let $B \to \beta_1 A
\beta_2$ be the only rule which contain $A$ in its right side, remove $A$ from
$V(G)$ and remove $A \to \alpha$ from $P(G)$, then replace the production rule
of $B$ by $B \to \beta_1 \alpha \beta_2$.

\item For an admissible grammar $G$, if there is a production rule $A \to
\alpha_1\beta\alpha_2\beta\alpha_3$ where $|\beta| > 1$, add a new non-terminal
$B$ into $V(G)$ then create a new rule $B \to \beta$, replace the production of
$A$ by $A \to \alpha_1 B \alpha_2 B \alpha_3$.

\item For an admissible grammar $G$, if there are two production rules $A_1$ and
$A_2$ that $A_1 \to \alpha_1\beta\alpha_2$ and $A_2 \to \alpha_3\beta\alpha_4$,
in witch $|\beta| > 1$ and either $|\alpha_1| > 0$ or $|\alpha_2| > 0$, either
$|\alpha_3| > 0$ or $|\alpha_4| > 0$, add a new non-terminal $B$ into $V(G)$ then
create a new rule $B \to \beta$, replace the production of $A_1$ by $A_1 \to
\alpha_1 B \alpha_2$, replace the production of $A_2$ by $A_2 \to \alpha_3 B
\alpha_4$.

\end{enumerate}

\begin{figure}
\centering
\footnotesize
\begin{lstlisting}
@\#Transform $x$ into an admissible grammar@
@\#returns the start rule by $p_0$, other rules by $G$@
def GrammarTransform($x$):
  $G$ = $\{\}$
  $p_0$ = SeqTransform($x$, $G$)
  return $p_0$, $G$

@\#$x$ is the sequence to be transform@
@\#$G$ is a set of production rules@
@\#output: return the start symbol $p_x$ so that $f_{G}^{\infty}(p_x) = x$,@
@\#        all other rules are added into $G$@
def SeqTransform($x$, $G$):
  $p_x$ = $S_x \to \epsilon$
  while $|x| > 0$:
    @\#greedy read ahead and match@
    for $p$ in $G$:
      $v$ = @left side of $p$@
      @check whether $f_{G}^{\infty}(v)$ is $x$'s prefix@
    if @matched@:
      $v$ = @the longest matched nonterminal@
      @append $v$ after the right side of $p_x$@
      @pop $|f_{G}^{\infty}(v)|$ entries from $x$@
    else:
      @pop one entry $t$ from $x$@
      @append $t$ as a terminal after the right side of $p_x$@
    @apply reduction rules 1-3 iteratively over $G \cup \{p_x\}$,@
    @until non of them can be applied.@
    @newly created rules are added into $G$@
  return $p_x$
\end{lstlisting}

\caption{Greedy grammar transform algorithm}\label{fig:greedyalgo}
\end{figure}

Figure \ref{fig:greedyalgo} illustrates the grammar transform
algorithm\cite{yang2000efficient}. It is
very similar to SEQUITUR\cite{nevill1997identifying, nevill1997compression}
except the greedy read ahead step, which guarantees that in the generated
grammar $G$, for different $v \in V(G)$, $f_{G}^{\infty}(v)$ are different.

Algorithm in figure \ref{fig:greedyalgo} transforms a sequence into a
context-free grammar.  To avoid patterns across different sequences interfering
the processing, we wrap the algorithm as figure \ref{fig:seqsalgo}. In the
wrapped algorithm, we can guarantee that every execution sequences are
represented by an non-terminal in the right side of $p_0$.
The reduction rules never consider patterns
across sequences because $p_0$ is not in $G$. In figure \ref{fig:seqsalgo} we also show
that our algorithm eliminates redundant sequences by dropping those results
which contain only one symbol.

\begin{figure}
\centering
\footnotesize
\begin{lstlisting}
@\# \texttt{logs} is a set of sequences@
def LogTransform(logs):
  $G$ = $\{\}$
  $p_0$ = $S_0 \to \epsilon$
  for $seq$ in logs:
    $p_n$ = SeqTransform($seq$, $G$)
    if @$p_n$ contains only one symbol in its right side@:
      @drop $p_n$@
      continue
    @insert $p_n$ into $G$@
    @append $p_n$ after the right side of $p_0$@
  return ($p_0$, $G$)
\end{lstlisting}
\caption{Transform a set of sequences}\label{fig:seqsalgo}
\end{figure}

In table \ref{tab:example} we explain the above algorithm using an example of
computing a grammar for 4 sequences \textsl{1234 1235 1234 1237}. The final
grammar is listed at the last row in the table.

\doublerulesep 0.1pt
\begin{table}
\begin{scriptsize}
\caption{Example of 4 sequences: 1234 1235 1234 1237}
\label{tab:example}
\begin{tabular}{llll}
\hline\hline\noalign{\smallskip}
processed & $p_n$ & $G$    &  $p_0$  \\
string    &       &        &       \\
\noalign{\smallskip}
\hline
\multicolumn{4}{l}{\bf begin process sequence 1234 } \\
     & $p_1 \to \epsilon$ & $\{\} $ & $p_0 \to \epsilon$ \\
\hline
1234 & $p_1 \to 1234$  &  &  \\
\hline
\multicolumn{4}{l}{\bf begin process a new sequence 1235} \\
     & $p_2 \to \epsilon$ & $\{p_1\}, p_1 \to 1234$ & $p_0 \to p_1$ \\
\hline
12   & $p_2 \to 12$    &                              & \\
\multicolumn{4}{l}{\bf apply rule 3 on pattern $12$} \\
     & $p_2 \to p_a$   & $\{p_1, p_a\}, p_1 \to p_a34$ & \\
     &                 & $\quad p_a \to 12$                 & \\
\hline
123  & $p_2 \to p_a3$  &                                    & \\
\multicolumn{4}{l}{\bf apply rule 3 on pattern $p_a3$} \\
     & $p_2 \to p_b$   & $\{p_1, p_a, p_b\}, p_1 \to p_b4$ & \\
     &                 & $\quad p_a \to 12, p_b \to p_a3$  & \\
\multicolumn{4}{l}{\bf apply rule 1 on $p_a$} \\
     &                 & $\{p_1, p_b\}, p_1 \to p_b4$      & \\
     &                 & $\quad p_b \to 123 $              & \\
\hline
1235 & $p_2 \to p_b5$  &                                   & \\
\hline
\multicolumn{4}{l}{\bf begin process a new sequence 1234} \\
     & $p_3 \to \epsilon$ & $\{p_1, p_b, p_2\}, p_1 \to p_b4$ & $p_0 \to p_1p_2$ \\
     &                    & $\quad p_b \to 123, p_2 \to p_b5$ & \\
\hline
\multicolumn{4}{l}{\bf look ahead greedy match: } \\
\multicolumn{4}{l}{$p_1$ and $p_b$ matched, $|f_{G}^{\infty}(p_1)|$ is the longest} \\
1234 & $p_3 \to p_1$   &                                   & \\
\hline
\multicolumn{4}{l}{\bf begin process a new sequence 1237} \\
\multicolumn{4}{l}{$p_3$ contains only 1 symbol, eliminate $p_3$} \\
     & $p_4 \to \epsilon$ &                                & $p_0 \to p_1p_2$ \\
\hline
\multicolumn{4}{l}{\bf look ahead greedy match: } \\
\multicolumn{4}{l}{$p_b$ matched} \\
123 & $p_4 \to p_b$    &                                   & \\
\hline
1237 & $p_4 \to p_b7$   &                                   & \\
\hline
\multicolumn{4}{l}{\bf finish processing } \\
     &                 & $\{p_1, p_b, p_2, p_4\}, p_1 \to p_b4$ & $p_0 \to p_1p_2p_4$ \\
     &                 & $\quad p_b \to 123, p_2 \to p_b5$  & \\
     &                 & $\quad p_4 \to p_b7$               & \\
\hline\hline
\end{tabular}
\end{scriptsize}
\end{table}

\begin{figure}
\centering
\footnotesize
\begin{lstlisting}
@\# Count the number of total symbols which is needed for@
@\# describing all rules in \texttt{rules}@
def EvaluateRules(rules, $G$):
  v = 0
  processedrules = $\{\}$
  for $r$ in rules:
    if @$r$ is in processedrules:@
      continue
    processedrules.insert($r$)
    @\#r is a production rule with the form $A \to \alpha$@
    v += $|\alpha|$
    for symbol in $\alpha$:
      if @symbol is nonterminal@:
        $r_n$ = $G$[symbol]
        if $r_n$ not in processedrules:
          rules.append($r_n$)
  return v

@\# $x$ is the sequence which is to be evaluated@
@\# $p_0$ and $G$ are parameters of an already computed grammar@
def EvaluateSequence($x$, $p_0$, $G$):
  $G'$ = $G$ @\#deep copy@
  $p_n$ = SeqTransform($x$, $G'$)
  info_old = EvaluateRules($\{p_0\}$, $G$)
  info_new = EvaluateRules($\{p_0, p_n\}$, $G'$)
  $I$ = info_new - info_old
  $D$ = $I$ / $|x|$
  return $I$, $D$
\end{lstlisting}
\caption{Evaluation of a new sequence}\label{fig:eval}
\end{figure}

We measure the quantities of information of a sequence by computing the number
of additional symbols which it introduces into the grammar.  Figure
\ref{fig:eval} describes the evaluating process. After a grammar generated,
\texttt{EvaluateSequence} is used to compute the information quantity ($I$) and
information density $D$ (average symbols produced by an entry) of a sequence
$x$. To illustrates the evaluating process, we evaluate sequences \textsl{2238}
and \textsl{1239} using $G$ generated by table \ref{tab:example}.

\begin{table}
\begin{footnotesize}
\caption{Evaluation of 2 sequences}
\begin{tabular}{p{2cm}p{2cm}p{2cm}}\small
$G$ & 2238 & 1239 \\
\hline
\begin{minipage}[t]{.3\textwidth}
$p_0 \to p_1p_2p_4$ \\
$p_b \to 123$ \\
$p_1 \to p_b4$ \\
$p_2 \to p_b5$ \\
$p_4 \to p_b7$ \\
\end{minipage} &
\begin{minipage}[t]{.3\textwidth}
$p_{2238} \to 2p_c8$ \\
$p_0 \to p_1p_2p_4$ \\
$p_b \to 1p_c$ \\
$p_c \to 23$ \\
$p_1 \to p_b4$ \\
$p_2 \to p_b5$ \\
$p_4 \to p_b7$ \\
\end{minipage} &
\begin{minipage}[t]{.3\textwidth}
$p_{1239} \to p_b9$ \\
$p_0 \to p_1p_2p_4$ \\
$p_b \to 123$ \\
$p_1 \to p_b4$ \\
$p_2 \to p_b5$ \\
$p_4 \to p_b7$ \\
\end{minipage} \\
\hline
12 symbols & 16 symbols & 14 symbols \\
	   & $I = 4$    & $I = 2$ \\
	   & $D = 1$    & $D = 0.5$ \\
\end{tabular}
\end{footnotesize}
\end{table}

From the above table, \textsl{2238} is more strange than \textsl{1239}.

\subsection{Anomaly detection based on compression}

We introduced out anomaly detection algorithm in this subsection.

The goal of the algorithm is to find abnormal sequences in given logs. The input
are two data sets. One set contains some normal sequences, the other set
contains questionable sequences. From the later set our algorithm reports
abnormal sequences.

The algorithm can be divided into following steps:
\begin{enumerate}

\item Training: transform the normal set $S_{n}$ into an admissible grammar $G$
with $S(G) = p_0$.

\item Evaluating: for each sequences $t_n$ in questionable set $S_{q}$, compute
$(I_{t_n}, D_{t_n})$ using \texttt{EvaluateSequence($t_n$, $p_0$, $G$)}.

\item Reporting: report $m_1$ sequences which generates largest $m_1$ $I_{t_n}$, 
report $m_2$ sequences which generates largest $m_2$ $D_{t_n}$. $m_1$ and $m_2$
are configurable.

\end{enumerate}

We report abnormal sequences according to both $I$ and $D$ because we believe
they are both meaningful. A sequence $x$ which generates large $I_x$ indicates
that there is no a similar sequence in $S_{n}$. However, if $x$ is a very long
sequence, the symbols used to describe $x$ may be at very high level. Compare
with a short sequence $y$ with $I_y \approxeq I_x$, $y$ is more valuable.

\section{Experimental analysis}\label{sect:experiment}

\subsection{Fine grained execution log}

We tested the ability of our method on finding bugs in fine grained execution
log. The data sets we used are generated using \texttt{ReBranch}\cite{rebranch}.
\texttt{ReBranch} is a record-replay tool for debugging. It records the outcome
of all branch instructions when running, and replay the execution according to
these logs for debugging. We converted the traces into line number sequences.
Figure \ref{fig:rebranchsample} shows a piece of sample trace.

\begin{figure}
\footnotesize
\begin{verbatim}
...
main:/../../server.c:516
main:/../../server.c:536
main:/../../server.c:538
server_init:/../../server.c:170
server_init:/../../server.c:172
...
\end{verbatim}
\caption{Sample \texttt{ReBranch} trace}\label{fig:rebranchsample}
\end{figure}

We tried our algorithm on finding two non-deterministic bugs in
\texttt{lighttpd} (a light weight web server) and \texttt{memcached} (a
key-value object caching system).

In \texttt{lighttpd} bug 2217\cite{lighttpdbug2217}, sometimes a few of CGI
requests timeout. The bug is caused by a race condition: when a child process
exits before the parent process is notified about the state of the corresponding
pipe, the parent will wrongly remove the pipe from the event pool and never
close the connection because it assumes the pipe still contain data.

In our experiment on \texttt{lighttpd} bug, we first collected a trace with 500
correct requests for training, then collected another trace with 1000 requests
for testing. 2 of these 1000 requests timeout. Traces are pre-processed to be
divided into sequences. During the pre-processing, signal handling are removed.
A sequence begin at the entry of \texttt{connection\_state\_machine()} and end
at the exit point of that function. After pre-processing, the normal trace
contains 2501 sequences made by 3337395 entries, the questionable trace contains
4996 sequences made by 6661425 entries.

\texttt{memcached} bug 106\cite{memcachedbug106} is combined by 2
bugs. We first fixed a udp deadlock problem under the help of \texttt{ReBranch}.
After that, when the cache server receives a magic udp packet, some of following
udp requests won't get reply. The problem is cause by incorrect state
transfer. \texttt{memcached} uses a state machine when serving a request.
The incorrect state transfer is \texttt{conn\_read -> conn\_closing}. The
correct transfer sequence is more complex.

In \texttt{memcached} experiment, we first collected a trace with 1000 correct
udp requests, then tried to identify 3 buggy requests out of 1003 new requests.
As previous experiment, we split traces into sequences. A sequence begin
at the entry point of \texttt{event\_handler()} and end at the exit point of
that function. After splitting, normal trace data set contains 1442 sequences
made by 862636 entries; questionable trace contains 1609 sequences made by
883226 entries.

\begin{table}
\begin{footnotesize}
\caption{Test result of \texttt{ReBranch} data sets}
\label{tab:rebranchresult}
\begin{tabular}{ll}
\hline\hline
\bf lighttpd & \rm train result: \\
             & 3337395 entries into 2793 symbols \\
\hline
\bf top most 5 $I$ & \bf top most 5 $D$ \\
\hline
$I_{654}  = 41$ & $D_{654}  = 0.039653$\\
$I_{3990} = 41$ & $D_{3990} = 0.039653$\\
$I_{1172} = 3 $ & $D_{655}  = 0.019231$\\
$I_{1}    = 1 $ & $D_{3991} = 0.019231$\\
$I_{2}    = 1 $ & $D_{22}   = 0.018868$\\
\hline
\bf memcached & \rm train result: \\
              & 862636 entries into 582 symbols \\
\hline
\bf top most 5 $I$ & \bf top most 5 $D$ \\
\hline
 $I_{1237} = 27$  & $D_{1237}  = 0.031765$\\
 $I_{1608} = 27$  & $D_{1608}  = 0.031765$\\
 $I_{1609} = 27 $ & $D_{1609}  = 0.031765$\\
 $I_{1}    = 1 $  & $D_{1}     = 0.009091$\\
 $I_{2}    = 1 $  & $D_{6}     = 0.009091$\\
\hline\hline
\end{tabular}
\end{footnotesize}
\end{table}

The results of the above 2 experiments are listed in table
\ref{tab:rebranchresult}. In \texttt{lighttpd} experiment, our algorithm find 2
sequences (654 and 3990) with $I$ and $D$  quite larger than others. In
\texttt{memcached} experiment, our algorithm find 3 strange sequences (1237,
1608 and 1609). We confirmed those sequences are correct ones (buggy ones) by
manually replaying.

It is hard to detect \texttt{memcached} 106 bug using traditional Markov-based
intrusion detection method because the misbehavior is at a very high level.
Markov-based methods only consider the probabilities of one entry transfer to
another entry. However, in this example, state transfer operation is implemented
by many lines, each line transfer is valid. If developer know the distance
between the key lines which represent a state transfer, higher order Markov
model or n-gram model can be used. Nevertheless, for different program,
developer have to manually adjust the length of sliding window. Furthermore,
computing higher order model requires much more resources-- always growths
exponentially.

\subsection{System call sequences}

We used the data set published by the University of New
Mexico\cite{warrender1999detecting} to evaluate the ability of our 
algorithm on intrusions detection. The published data sets are system
call traces generated using \texttt{strace}. 

We applied our algorithm on \texttt{xlock} and \texttt{named} data sets. Figure
\ref{fig:unmdata} shows the size of those data and some sample entries in those
traces. A trace entry contains two numbers, the left one is process id, the
right one is the system call number. A trace in UNM data set contains many
processes.

\begin{figure}
\footnotesize
\begin{minipage}[c]{7cm}
\begin{tabular}{llll}
\hline\hline
data set & traces & procs & entries \\
\hline
xlock-synth-unm & 71 & 71 & 339177 \\
xlock-intrusions & 2 & 2  & 949 \\
\hline
named-live & 1 & 27 & 9230572 \\
named-exploit & 2 & 5 & 1800 \\
\hline\hline
\end{tabular}
\end{minipage}%
\begin{minipage}[c]{2cm}
\begin{verbatim}
...
229 2
229 1
370 66
370 5
370 63
...
\end{verbatim}
\end{minipage}
\caption{UNM data set sample and size}\label{fig:unmdata}
\end{figure}

We use our algorithm to identify exploited processes. To achieve this, we
splitted the original traces into system call sequences according to process id.
The entries in each result sequences contain only the system call number.  For
\texttt{xlock}, we randomly selected 61 processes sequences for training then
compared $I$ and $D$ of the other 12 sequences (10 normal, 2 exploited); for
\texttt{named}, we chose 22 of normal sequences for training. The result is listed
in table \ref{tab:unmtest}.

\begin{table}
\begin{footnotesize}
\caption{Test result of UNM data sets}
\label{tab:unmtest}
\begin{tabular}{llll}
\hline\hline
\bf xlock & \multicolumn{3}{l}{\rm train result: 266563 entries into 6149 symbols} \\
\hline
\multicolumn{2}{c}{top most 5 normal sequences} &
\multicolumn{2}{c}{2 exploited sequences}\\
\hline
$I_5 = 776$ & $D_{10} = 0.107226$ & $I_{q_1} = 183$ & $D_{q_1} = 0.372709 $\\
$I_8 = 89$ & $D_1 = 0.049743$ & $I_{q_2} = 176$ & $D_{q_2} = 0.380952 $ \\
$I_1 = 87$ & $D_5 = 0.036998$ & & \\
$I_4 = 59$ & $D_6 = 0.032590$ & & \\
$I_{10} = 46$ & $D_4 = 0.026375$ & & \\
\hline
\bf named & \multicolumn{3}{l}{\rm train result: 9215497 entries into 66148 symbols} \\
\hline
\multicolumn{2}{c}{top most 5 normal sequences} &
\multicolumn{2}{c}{5 exploited sequences}\\
\hline
$I_2 = 322$ & $D_2 = 0.031078$ & $I_{q_2} = 90$ & $D_{q_3} = 0.311688$ \\
$I_3 = 4$ & $D_3 = 0.003537$   & $I_{q_5} = 70$ & $D_{q_2} = 0.297030$ \\
$I_4 = 4$ & $D_5 = 0.003537$   & $I_{q_3} = 24$ & $D_{q_5} = 0.291667$ \\
$I_5 = 4$ & $D_4 = 0.002093$   & $I_{q_1} = 1 $ & $D_{q_1} = 0.001681$ \\
$I_1 = 1$ & $D_1 = 0.001681$   & $I_{q_4} = 1 $ & $D_{q_4} = 0.001681$ \\
\hline\hline
\end{tabular}
\end{footnotesize}
\end{table}

In \texttt{xlock} result, information density ($D$) of the two exploited
sequences are 2 times larger than the largest density in normal set.  In
\texttt{named} result, our algorithm identified 3 strange sequences, information
densities of them are at different order of magnitude. The last 2 sequences
generate only 1 symbol ($I = 1$), indicates that same sequences have appeared in the
training set at least once. After checking we found that those 2 processes are
the parent processes used to setup daemons, none of them is target of attacks.

\subsection{Performance}

Finally we list the throughput of our algorithm in table \ref{tab:speed}. It has
been shown that SEQUITUR is a linear-time algorithm\cite{nevill1997identifying}.
Our algorithm is similar to SEQUITUR except the read ahead matching. Such
matching (match a long string against many shorter strings and find the longest
match) can be optimized using a prefix tree.

\begin{table}
\begin{footnotesize}
\caption{Processing speed}
\label{tab:speed}
\begin{tabular}{lcccc}
\hline\hline
data set & \multicolumn{2}{c}{training} & \multicolumn{2}{c}{evaluating} \\
         & time & throughput & time & throughput \\
	 & ($s$) & ($ent/s$) & ($s$) & ($ent/s$) \\
\hline
\texttt{lighttpd} & 90.4 & 36918.1  & 496.3 & 13422.2 \\
\texttt{memcached} & 10.1 &  85409.5  & 28.7 & 30817.4 \\
\texttt{xlock} & 237.1 & 1124.3  & 166.7 & 442.1 \\
\texttt{named} & 15742.6 & 585.4 & 189.9 & 89.2 \\
\hline\hline
\end{tabular}
\end{footnotesize}
\end{table}

\section{Conclusion}\label{sect:conclusion}

In this paper we propose a novel anomaly detection algorithm by comparing the
incrementation of compressed data length based on grammar-based compression. To
the best of our knowledge, this is the first work which uses compression to
measure the strangeness of sequences in anomaly detection. Different from
Markov-based algorithm, our method utilizes the full knowledge about the
structure of the data set. It can be used to find high level misbehavior as well
as low level intrusions.  We tested the algorithm on finding bugs in fine
grained execution logs and intrusion detection in system call traces. In both
data set, our method got positive result. The proposed method is also applicable
to text log generated by today's cloud computing systems.

\begin{small}
\noindent {\bf Acknowledgements}\quad This work is partially supported by the National Natural
Science Foundation of China (Grant No. 61070028 and 61003063).
\end{small}

\bibliographystyle{unsrt}
\bibliography{oo}

\end{document}